\documentclass[runningheads]{llncs}
\usepackage[T1]{fontenc}
\usepackage{graphicx,verbatim}
\usepackage{xcolor}
\usepackage{amsmath}
\usepackage{amssymb}
\usepackage{hyperref}
\usepackage{booktabs}
\usepackage{multirow}
\usepackage{color}

\begin{document}
\title{OTLesMix: Wasserstein Barycenter and Optimal Transport Map for Synthetic Lesion Generation with Diverse Shapes and Locations}
\titlerunning{OTLesMix: Synthetic Lesion Generation with Optimal Transport}

\author{Robin Trombetta\inst{}\orcidID{0009-0001-7945-6953} \and
Carole Lartizien\inst{}\orcidID{0000-0001-7594-4231}
}
\authorrunning{R. Trombetta et al.}

\institute{Univ. Lyon, CNRS UMR 5220, Inserm U1294, INSA Lyon, UCBL, CREATIS, France \email{robin.trombetta@creatis.insa-lyon.fr}
}

\maketitle              
\begin{abstract}
The development of deep learning over the past decade has revolutionized medical imaging segmentation, allowing the extraction of precise descriptors from large volumes to characterize pathologies. 
Data augmentation is a technique widely regarded as a way to improve model training. It includes simple transformations like spatial operations or intensity modifications, but also more advanced synthesis techniques. Their goal is to generate new realistic samples from an existing dataset to diversify the images used during training. Among them, several propose different mixing strategies to combine real samples. However, one of their major shortcomings is to yield limited variability in terms of generated lesion shapes and locations. In this work, we introduce a novel image synthesis method, called OTLesMix, that leverages Wasserstein barycenter and optimal transport plan to generate realistic and diverse samples. We evaluated our method on three brain lesion segmentation tasks, on which it improves the Dice score compared to a model trained without synthetic data by 2.9 to 6.6 points, and outperforms state-of-the-art \textit{mix-based} methods.


\end{abstract}

\section{Introduction}

Deep learning models for image segmentation have opened new perspectives for medical image analysis, allowing for a fine description of anatomical or functional structures. Supervised training remains the dominant and most performing training paradigm, but the cost of getting annotations can be prohibitive, especially for medical imaging applications. It requires expert knowledge and is time-consuming, especially for tasks like dense image segmentation which can take up to several hours for a single 3D imaging volume \cite{dang2022vessel}.
 
Data augmentation is a broadly used technique to improve the training of neural networks. It works by increasing the diversity of the training samples
through the application of some transformations to the available data. Basic random image transformations are spatial and intensity transformations, or noise injection.

To improve further the diversity of generated samples, more advanced techniques leveraging generative models, like GANs or Diffusion Models, have also been widely studied over the past few years. Such methods yield state-of-the-art results in terms of realism of the generated samples. Conditioning the generation process on additional attributes, such as class, mask or text, also allows guiding the sampling process \cite{dorjsembe2024conditional,zhang2024diffboost}. 
However, such generative models demand prior training based on diverse and numerous data, which can be a deceptive solution. If the available data is restricted, training a generative deep learning model may not be efficient for generating new realistic data.

To overcome the limitations of methods based on deep generative models, other approaches propose to mix available annotated data with simpler operations. Two early approaches falling into this category of \textit{mix-based} methods are MixUp \cite{zhang2018mixup} and CutMix \cite{yun2019_cutmix}. In the former, a new sample is produced by a random linear combination of two existing images and their associated ground truth. The latter consists in cutting a rectangle in a first image and pasting it in a second image to corrupt it. These two methods have been primarily designed for classification tasks. 

Some works focus on designing data augmentation techniques more suited to segmentation tasks, and in particular for medical imaging applications. CarveMix \cite{ZHANG2023_carvmix} proposes to carve an image and its segmentation mask according to the location and geometry of the ground truth lesion, and inpaint the resulting region of interest onto another image. LesionMix \cite{Berke2023_lesionmix} generates lesions using a marching method to augment existing images, allowing the creation of more diverse lesion shapes, with control over the synthetic lesion load. SelfMix \cite{zhu2022_selfmix} fuses several random lesion and non-lesion regions from the training dataset with computed weights to create new samples. These data augmentation techniques can produce unrealistic synthetic images, in particular when compared to samples generated with GANs or Diffusion Models, but recent works have shown that such \textit{out-of-domain} images can still improve performance of deep segmentation models \cite{cubuk2021_tradeoff_DA}. There is a tradeoff between the domain gap between real and synthetic images, and the diversity of generated samples. 

As can be seen in the most recent research works in the field of medical imaging data augmentation, the main areas of improvement are the realism of generated images and the diversity of the shapes and locations of the masks. The new shapes are usually obtained via a simple random sampling process from existing shapes \cite{Berke2023_lesionmix,zhu2022_selfmix}, which gives limited diversity. Similarly, the locations of the generated lesion masks are drawn from existing ones \cite{ZHANG2023_carvmix,Berke2023_lesionmix}, sometimes with additional simple transformations, such as rotations or translations \cite{zhu2022_selfmix}, which restrict their variety.

In this work, we propose to leverage Optimal Transport (OT) to produce synthetic samples with high diversity both in terms of lesion shapes and locations. Our method consists of two main steps. First, given two or more lesion images from the training dataset with their respective masks, referred to as base images in the following,
we compute the Wasserstein barycenter of the masks to generate a new lesion shape and location. The obtained mask can be located anywhere between the two original masks, and its shape is the interpolation between the base masks. Since the generated mask does not fully overlap any anomalous region in the base images, we then compute the optimal transport plan between the generated mask and one of the two base masks and use it to derive the intensities of the generated image inside the lesion mask. The synthesized lesion is inpainted onto one of the two base images, or onto any other image from the training dataset. We evaluate our approach, dubbed OTLesMix, on three brain segmentation tasks, on which it is shown to outperform state-of-the-art \textit{mix-based} data augmentation techniques. The code that implements our method and reproduces the experiments of this study is available at \href{https://github.com/robintrmbtt/otlesmix}{https://github.com/robintrmbtt/otlesmix}.

\section{Method}
\label{sec:method}

Let us consider a dataset $\mathcal{D} = \{ (\mathbf{X}_i, \textbf{Y}_i)\}_{i=1}^n$, composed of pairs of images $\textbf{X}_i$ and associated pixel-level ground truth annotations $\textbf{Y}_i$, where $\textbf{Y}_i$ can take values in $\mathbb{N}$ in the case of a multi-class segmentation task. The goal is to generate a synthetic dataset $\mathcal{D_{\mathrm{synth}}} = \{(\mathbf{\tilde
{X}}_i, \mathbf{\tilde{Y}}_i)\}_{i=1}^m = f(\mathcal{D})$ from the base dataset $\mathcal{D}$. We hope that our data generation process $f$ is such that a model trained on $\mathcal{D} \cup \mathcal{D}_{\mathrm{synth}}$ achieves better performance on a test set than a baseline model trained only on $\mathcal{D}$.

\subsection{Overview of the method}

\begin{figure}[t]
\centering
\includegraphics[width=\textwidth]{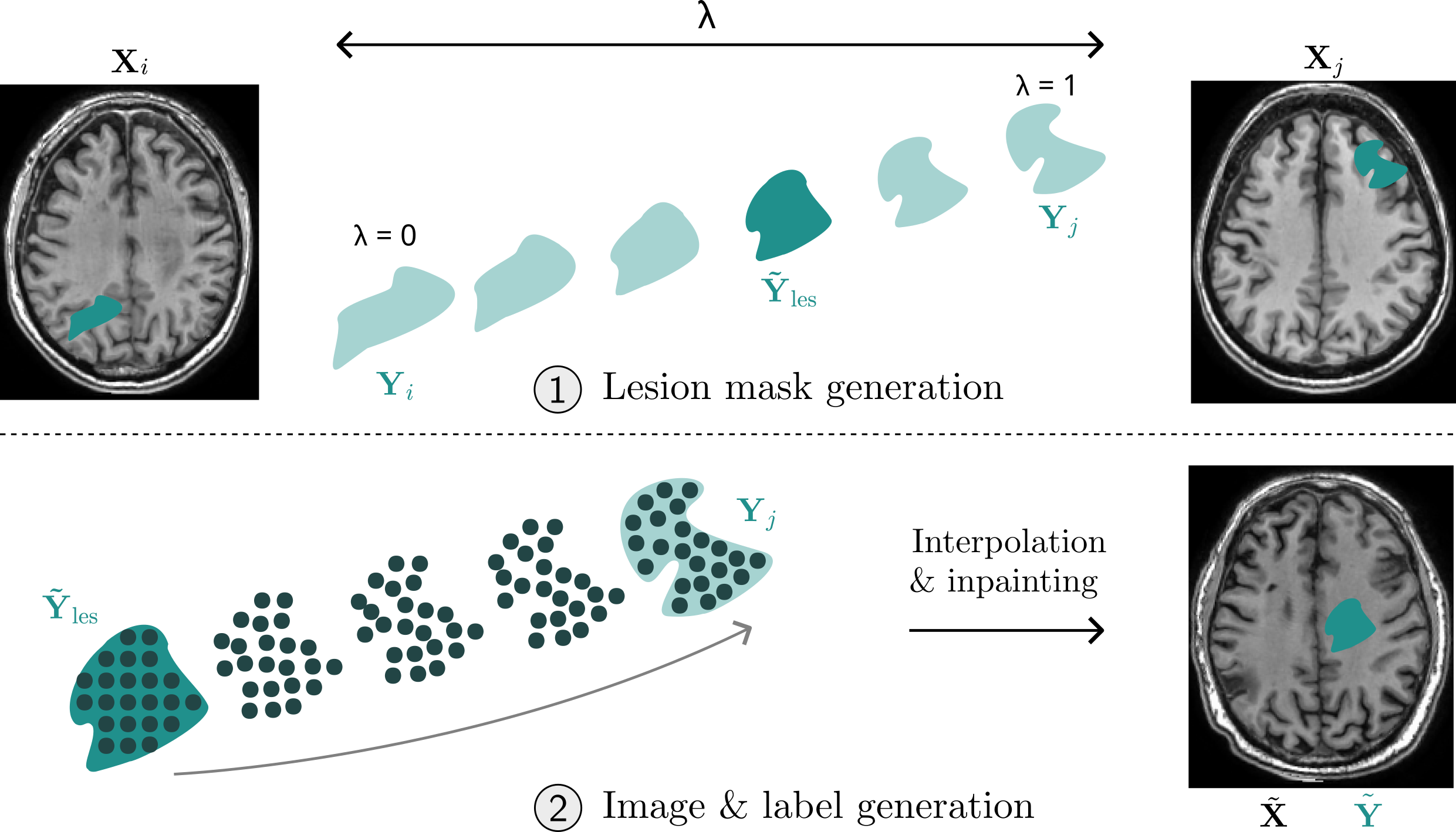}
\caption{Illustration of the sample generation process with OTLesMix.} 
\label{fig:illustration}
\end{figure}

Our method, illustrated in Figure \ref{fig:illustration}, is composed of two main steps, described in more detail in Sections \ref{subsec:lesion_mask_generation} and \ref{subsec:image_label_generation}. The first step consists in combining two ground truth masks to create a new lesion mask $\mathbf{\tilde{Y}}_{\mathrm{les}}$. We randomly draw two base masks $\textbf{Y}_i$ and $\textbf{Y}_j$ from the training dataset and compute their Wasserstein barycenter to obtain $\mathbf{\tilde{Y}}_{\mathrm{les}}$. As shown in Figure \ref{fig:illustration}, this step allows interpolating between two existing shapes to generate original and realistic lesion shapes. Moreover, the location of the new lesion falls between the two base masks.

Once we have obtained a new lesion $\mathbf{\tilde{Y}}_{\mathrm{les}}$, the second step consists in computing both the intensities and the labels of its constituting pixels. To do so, we \textit{transport} every pixel of the lesion mask $\mathbf{\tilde{Y}}_{\mathrm{les}}$ onto either the lesion of $\mathbf{Y}_i$ or the lesion of $\mathbf{Y}_j$. For each pixel of the lesion mask $\mathbf{\tilde{Y}}_{\mathrm{les}}$, we look at where it is mapped by the transport plane, and interpolate the image intensity values and the label accordingly. Finally, we can inpaint the synthetic lesion onto $(\mathbf{X}_i, \textbf{Y}_i)$, $(\mathbf{X}_j, \textbf{Y}_j)$ or any other sample from the training dataset to generate a new sample $(\tilde{\mathbf{X}}, \tilde{\textbf{Y}})$.

\subsection{Lesion mask generation with Wasserstein barycenter}

\label{subsec:lesion_mask_generation}

Let us take two samples $(\mathbf{X}_i, \textbf{Y}_i)$ and $(\mathbf{X}_j, \textbf{Y}_j)$ from the training dataset. We denote by $\alpha_i \overset{def}{=}(x_{i,k}) \in \mathbb{R}^{N_i \times D}$ and $\alpha_j \overset{def}{=}(x_{j,k}) \in \mathbb{R}^{N_j \times D}$ the coordinates of the $N_i$ (resp. $N_j)$ points of dimension $D$ where the respective label maps $\textbf{Y}_i$ (resp. $\textbf{Y}_j$) are not null. To interpolate between the two lesion shapes of $\textbf{Y}_i$ and $\textbf{Y}_j$, we make use of the theory of optimal transport, which provides a means to compare two probabilistic distributions. More formally, if we consider the distance, or \textit{cost}, between two points $\mathrm{C} : (x,y) \in 
\mathbb{R}^{D} \times \mathbb{R}^{D} \rightarrow \mathbf{C}(x,y) = \frac{1}{2}\lVert x - y \rVert^2 \in \mathbb{R}_{\geq 0}$, Kantorovich's formulation of the optimal transport problem consists in finding the transport plan $\pi \in \mathbb{R}_{\geq 0}^{N_i \times N_j}$ that minimizes the loss:

\begin{equation}
\label{eq:OT}
    \mathrm{OT}(\alpha, \beta) \overset{\mathrm{def}}{=} \min_{\pi\in \mathbb{R}_{\geq 0}^{N_i \times N_j}} \sum_{m=1}^{N_i} \sum_{n=1}^{N_j} \pi_{m,n} \mathbf{C}(x_{i,m}, x_{j,n})
\end{equation}
under the marginal constraints $\forall i,j, \sum_{n=1}^{N_j} \pi_{i,n} = \frac{1}{N_i}$ and $\sum_{m=1}^{N_i} \pi_{m,j} = \frac{1}{N_j}$. $\pi$ represents how the \textit{mass} of the points $\alpha_i$ is distributed among the points $\alpha_j$.

We denote by $\beta \overset{def}{=}(y_k) \in \mathbb{R}^{N \times D}$ the coordinates of the points of the new lesion mask, and consider a weighting parameter $\lambda$ randomly drawn in the distribution $\mathcal{U}(0,1)$. We optimize the coordinates of these points such that they are the Wasserstein barycenter of $\alpha_i$ and $\alpha_j$, \textit{i.e.} we minimize the loss:
\begin{equation}
    \mathrm{B}(\lambda, \alpha_i, \alpha_j, \beta) \overset{def}{=} \lambda \mathrm{OT}(\alpha_i, \beta) + (1 - \lambda) \mathrm{OT}(\alpha_j, \beta)
\end{equation}

When the optimal $\{y_k\}_{k=1}^N$ are found, we simply round their coordinates to the closest integer to obtain the final binary lesion mask $\tilde{\textbf{Y}}_{\mathrm{les}}$.

In practice, solving Equation (\ref{eq:OT}) does not scale well with the number of points $N_i + N_j$. We thus consider a regularized version of the OT cost, called \textit{de-biased Sinkhorn divergence} \cite{feydy2019interpolating}. This formulation can be solved fast using the Sinkhorn algorithm, with a linear instead of quadratic memory footprint, and for measures with up to $N \sim 1\mathrm{e}^{6}$ points in a matter of seconds on modern GPUs. 

\subsection{Image and label generation with optimal transport map}
\label{subsec:image_label_generation}

Given the coordinates $(y_k)$ of the points of the generated lesion mask $\tilde{\textbf{Y}}_{\mathrm{les}}$, we now want to \textit{transport} these points to either the lesion of $\mathbf{Y}_j$ -- or similarly $\mathbf{Y}_i$. More formally, we aim to find the optimal transport map $\mathrm{T} : x \in \mathbb{R}^{D} \rightarrow T(x) \in \mathbb{R}^{D}$ which minimizes $\sum_{k=1}^{N} \sum_{n=1}^{N_j} \mathbf{C}(\mathrm{T}(y_k), x_{j,n})$.

To obtain the intensities of the pixel of the synthetic lesion, for each point of the mask $\tilde{\textbf{Y}}_{\mathrm{les}}$ with coordinates $y_k$, we compute the linear interpolation of the point $\mathrm{T}(y_k)$ based on the intensities of the image $\mathbf{X}_j$. If the segmentation task is multi-class, we also compute the ground truth label map of the synthetic by nearest-neighbour interpolation of the point $\mathrm{T}(y_k)$ based on the ground truth label map $\mathbf{Y}_j$. Finally, the obtained lesion can be inpainted onto $(\mathbf{X}_i, \textbf{Y}_i)$, $(\mathbf{X}_j, \textbf{Y}_j)$ or any other sample from the training dataset to generate a new sample $(\tilde{\mathbf{X}}, \tilde{\textbf{Y}})$

\section{Experiments}

\subsection{Experimental Setup}

To evaluate the effectiveness of OTLesMix as a data augmentation technique to improve the training of a deep learning model, we apply the generation procedure described in Section \ref{sec:method} to increase the training datasets of several CNN-based segmentation tasks, both in 2D and 3D. For a given training dataset containing images and associated ground truth label maps, we repeatedly apply OTLesMix to pairs of training samples to generate synthetic data and increase the number of training samples. Then, two identical neural networks are trained, the first on real data only, the second on both real and synthetic data. Both models are evaluated solely on real-world data to assess the impact of using partially synthesized data during training and determine whether or not this improves the downstream segmentation performance.

\subsubsection{Datasets} We use three publicly available datasets for the segmentation of brain pathologies:
\begin{itemize}
    \item \textbf{BraTS 2020} \cite{menze2014brats,bakas2017brats} is a multi-institutional database of pre-operative MRI scans of glioblastoma and low-grade glioma. It contains images from  369 patients with four modalities: native T1-weighted, post-contrast T1-weighted, T2-weighted and FLAIR, and annotated by expert radiologists with three labels: enhancing tumor, the peritumoral edema, and the necrotic and non-enhancing tumor core. The volumes are provided already pre-processed, namely with co-registration, skull-stripping and resampling to isotropic resolution of 1mm\textsuperscript{3}. We extract 200 2D slices from 37 patients for the training, 100 with tumors and 100 without, and 10 000 slices with tumors from the remaining patients for the evaluation. On this dataset, we augment the training dataset by generating 1000 synthetic images.
    We create synthetic lesions from pairs of slices with tumors and inpaint them on non-tumorous slices.
    \item \textbf{ATLAS v2.0} \cite{liew2022atlasv2}. We leverage the 655 openly available T1w volumes of this manually annotated stroke neuroimaging dataset. The images are defaced, registered on the standard atlas MNI, and their intensities are normalized. In addition, we crop the volumes to keep only the 60 central slices, which gives us 3D images of size 197$\times$233$\times$60 mm\textsuperscript{3}. One-hundred 3D images are used for the training and the rest are reserved for evaluating the models. We synthesize 200 volumes on this dataset, by generating 3D synthetic lesions from pairs of 3D base volumes and inpainting them on either one or the other base volumes.
    \item \textbf{ISLES 2022} \cite{hernandez2022isles} is a challenge for Ischemic Stroke Lesion Segmentation, comprising 250 publicly available MRI cases of acute to sub-acute stroke lesions. Each patient has at least a 3D FLAIR and a Diffusion-Weighted Image (DWI) exam, but we only keep the latter modality in our experiments. The volumes are resampled to 2 mm\textsuperscript{3}, skull-stripped and registered to MNI-152 space. We use 50 volumes for the training, 200 for the evaluation, and generate 100 additional synthetic images, following the same protocol as for the ATLAS dataset.
\end{itemize}

\subsubsection{Compared methods} OTLesMix is compared to related state-of-the-art methods, namely MixUp \cite{zhang2018mixup}, CutMix \cite{yun2019_cutmix} and CarveMix \cite{ZHANG2023_carvmix}. We generate the same number of synthetic images and use the same random seed for all methods. Each training dataset, endowed with the synthetic data, is used within the framework nnUNet \cite{isensee2021nnunet} for fair comparison. nnUNet was chosen as one of the highest-performing state-of-the-art versatile backbone for medical image segmentation. 
In addition to synthetic data, all methods also use \textit{traditional data augmentations} (TDA) -- rotation, scaling, blurring, etc. --, which are included in the framework of nnUNet. As a baseline, we also report the performance of nnUNet trained without any synthetic data, \textit{i.e.} only with TDA.

\subsubsection{Implementation details} The Python package GeomLoss\footnote{\href{https://github.com/jeanfeydy/geomloss}{https://github.com/jeanfeydy/geomloss}} is used to implement our method, in particular to compute the Wasserstein barycenter and the optimal transport map described in Section \ref{sec:method}. The entropic penalty $\epsilon$ in Equation \ref{eq:OT} is set to $0.01$. To smooth the contours of the synthetic lesions, we apply a Gaussian filter with an isotropic kernel size of 1 mm. Moreover, when a synthetic lesion is inpainted on an image, we remove the parts of the lesion that intersect existing lesions or that land on unrealistic locations, such as the skull or the cerebrospinal fluid. Those post-processing steps are applied to all methods. The nnUNet models are trained for 1000 epochs on BraTS and 500 on ATLAS and ISLES. By default, the batch size is capped by the number of training samples, so it is smaller for the model trained without synthetic data. For fair comparison, we manually set the same batch size for all methods.

\subsubsection{Evaluation}

To evaluate the compared methods, we report the Dice Similarity Coefficient (DSC) for all datasets. On BraTS 2020, we follow the standard practice and report the DSC of 3 classes: Whole Tumor (WT), Tumor Core (TC) and Enhanced Core (EC). The statistical significance of the difference between the best-performing method and the others is assessed with a pairwise Wilcoxon signed-rank test.

\subsection{Results}

\begin{figure}[t!]
\centering
\includegraphics[width=\textwidth]{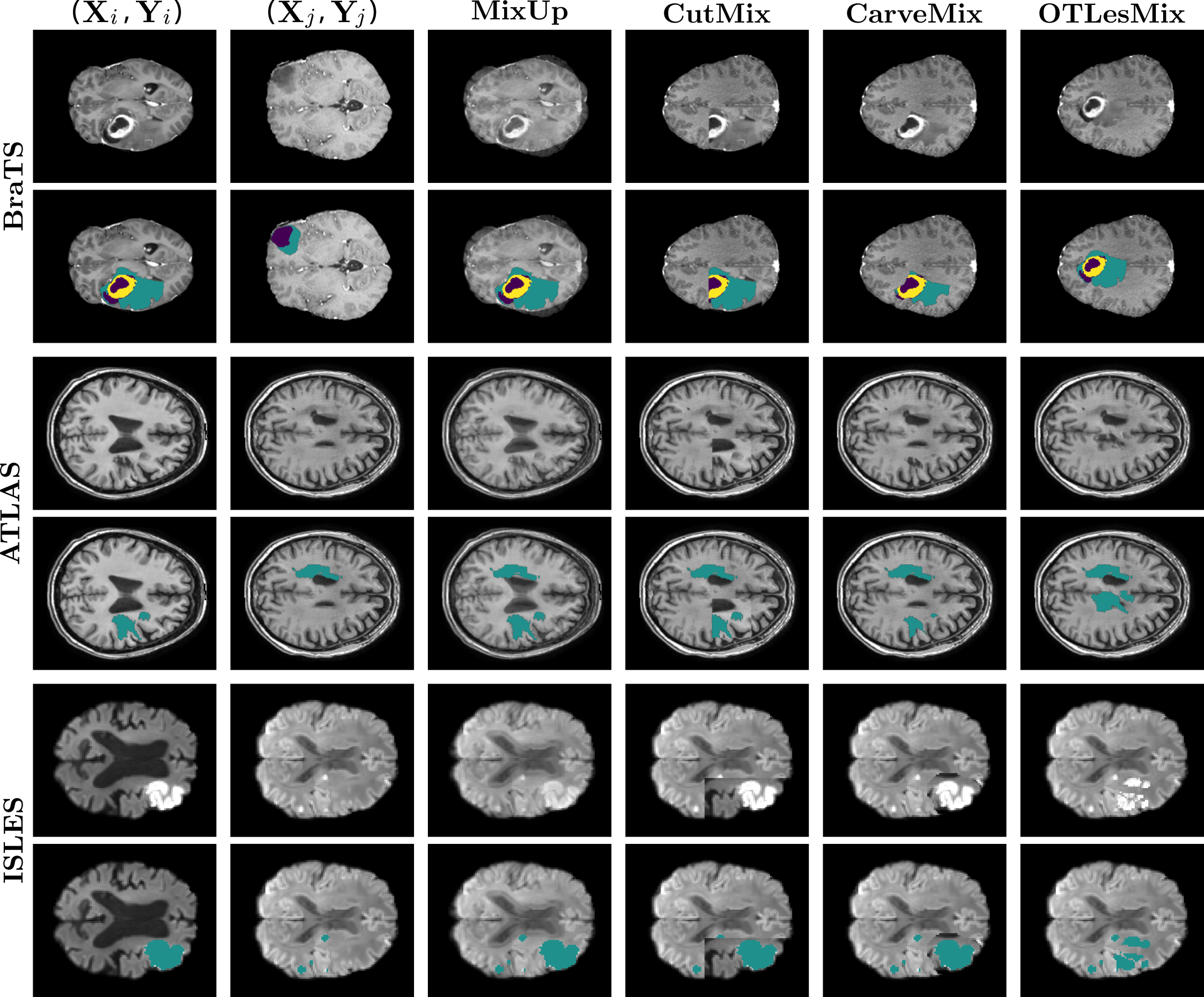}
\caption{Examples of generated samples from all methods. For each database, the first row shows the image and the second the corresponding ground truth label map. The first two columns represent the two base images with their corresponding original mask, while the other columns depict the resulting synthetic lesions inpainted on an non-tumorous 2D slice for BraTS and on either one or the other base 3D images for ATLAS and ISLES. Note that for BraTS, the synthetic tumors are inpainted on a third non-tumorous image, which is not shown here, and only OTLesMix uses $\mathbf{Y}_j$.} 
\label{fig:examples}
\end{figure}

Table \ref{table:results} presents the quantitative results on the three datasets and Figure \ref{fig:examples} shows examples of generated samples for all methods. 

On every dataset, OTLesMix outperforms all concurrent methods, increasing DSC by 0.8 to 1.8 points over the second-best method. Notably, the difference in DSC between OTLesmix and the other generative methods is statistically significant with a p-value of at least $0.05$, the greatest p-value being of $0.012$ for CarveMix on the ATLAS database. Compared to the standard nnUNet without data synthesis (TDA), our approach improves the DSC on the test set by between 2.9 points on ISLES and 6.6 points on the class WT of the BraTS dataset. 

Qualitatively, Figure \ref{fig:examples} illustrates how OTLesMix allows generating lesion masks with more diversity in shapes and locations than other state-of-the-art \textit{mix-based} methods. The example on the BraTS dataset -- first two rows -- shows that our method works on a multi-class segmentation setup. On ATLAS and ISLES, we can see that the shape interpolation computed with Wasserstein barycenter is consistent when the original lesion masks $\mathbf{Y}_i$ and $\mathbf{Y}_j$ contain multiple connected components.

\renewcommand{\arraystretch}{1.3}
\begin{table}[t]
\footnotesize
\caption{Average and standard deviation of the DSC (\%) on the three test datasets. The best model on each task is highlighted in \textbf{bold}. Asterisks indicate the level of statistical significance of the difference in DSC between a given method and the best performing one with a pairwise Wilcoxon signed-rank test ($^{*}$: $p \leq 0.05$, $^{**}$: $p \leq 0.01$, $^{***}$: $p \leq 0.001$).}

\begin{tabular}{@{}l@{\hspace{0.4cm}}lllll@{}}
\toprule
\multicolumn{1}{c}{} &
  \multicolumn{3}{c}{BraTS} &
  \multicolumn{1}{c}{\multirow{2}{*}{ATLAS}} &
  \multicolumn{1}{c}{\multirow{2}{*}{ISLES}} \\ \cmidrule(lr){2-4}
\multicolumn{1}{c}{} &
  \multicolumn{1}{c}{WT} &
  \multicolumn{1}{c}{TC} &
  \multicolumn{1}{c}{ET} &
  \multicolumn{1}{c}{} &
  \multicolumn{1}{c}{} \\ \midrule
TDA             & $77.0\pm 30.1^{***}$  & $72.5\pm29.6^{***}$ & $66.4\pm30.9^{***}$ & $45.8\pm33.8^{***}$ & $68.8\pm22.4^{***}$  \\
MixUp           & $80.6 \pm 24.5^{***}$ & $75.2 \pm 24.7^{***}$ & $68.2 \pm 27.6^{***}$ &  $43.7 \pm 33.0^{***}$  & $66.3 \pm 23.9^{***}$   \\
CutMix          & $81.9 \pm 22.5^{***}$ & $76.7 \pm 23.1^{***}$ & $70.0 \pm 26.7^{***}$ & $48.1 \pm 33.2^{**}$ & $69.8 \pm 22.6^{***}$  \\
CarveMix        & $82.5 \pm 22.3^{***}$ & $77.2 \pm 23.1^{***}$ & $69.6 \pm 26.9^{***}$ & $48.5 \pm 32.5^{*}$ & $70.8 \pm 21.4^{**}$ \\
OTLesMix        & $\mathbf{83.3 \pm 22.8}$ & $\mathbf{78.4 \pm 23.2}$ & $\mathbf{71.4 \pm 26.5}$ & $\mathbf{49.9 \pm 32.9}$ & $\mathbf{71.7 \pm 20.8}$  \\ \bottomrule
\end{tabular}
\label{table:results}
\end{table}

\section{Discussion and Conclusion}

In this work, we propose to leverage Wasserstein barycenter and optimal transport map to generate synthetic lesions and enhance the training of deep learning segmentation models. Compared to related methods that also mix real samples, our method yields more varied shapes and locations. Although it achieves better segmentation performance than the compared methods, OTLesMix still suffers from several limitations. 
Typically, generating one sample takes around 1 second for 2D images (BraTS) and 1-3 minutes on 3D datasets (ATLAS and ISLES), depending on the resolution and size of the images. 
Moreover, synthetic lesions are obtained simply by linear interpolation of the pixel intensities of real lesions. This approach can result in unrealistic inpainted lesions, in particular when there is a domain gap -- different scanner or acquisition protocol for instance -- between the two original base images $\mathbf{X}_i$ and $\mathbf{X}_j$. One perspective of this work would be to improve the quality of the synthetic samples by combining the mask generation procedure of OTLesMix with mask-conditioned deep-learning-based generation models.

    

\begin{credits}
\subsubsection{\ackname} 
This work was funded by the \textit{Agence Nationale de la Recherche} (ANR) under projects ANR-24-CE45-4399 (SEIZURE), ANR-11-INBS-0006 (FLI) and 11-LABX-0063 (Labex PRIMES).
This work was granted access to the HPC resources of IDRIS under the allocation 2026-AD011014900R2 made by GENCI.

\subsubsection{\discintname}
The authors have no competing interests to declare that are
relevant to the content of this article.
\end{credits}


%
%
%
\bibliographystyle{splncs04}
\bibliography{bibliography}
\end{document}